\documentclass{article}
\usepackage[english]{babel}
\usepackage{amsmath}
\usepackage{amssymb}

\usepackage{tikz}
\usepackage{pgfplots}

\usepackage{mathtools}
\usepackage[margin=1in]{geometry}
\usepackage[font=small]{caption}
\usepackage{url}
\usepackage{wrapfig}
\usepackage{soul}
\usepackage[font=footnotesize,labelfont=bf, labelsep=endash]{caption}
\usepackage{multirow}

\usepackage{todonotes}

\usepackage{chngcntr}
\usepackage{apptools}
\usepackage{paralist}

\usepackage{diagbox}
\usepackage{siunitx}
\usepackage{mathtools}
\usepackage{bbm}
\usepackage{algorithm}
\usepackage{algorithmic}
\usepackage{authblk}

\usepackage{array}
\definecolor{mathematicablue}{RGB}{94,129,181}
\definecolor{mathematicayellow}{RGB}{231,176,80}

\newcommand{\bnb}[1]{{B\&B}}

\colorlet{intenseorange}{mathematicayellow!90}
\colorlet{mediocreorange}{mathematicayellow!55}
\colorlet{fairorange}{mathematicayellow!25}
\DeclareRobustCommand{\hlintense}[1]{{\sethlcolor{intenseorange}\hl{#1}}}
\DeclareRobustCommand{\hlmediocre}[1]{{\sethlcolor{mediocreorange}\hl{#1}}}
\DeclareRobustCommand{\hlfair}[1]{{\sethlcolor{fairorange}\hl{#1}}}

\title{Learning to Schedule Heuristics in Branch-and-Bound\footnote{This work 
was partially funded by the Deutsche Forschungsgemeinschaft (DFG, German 
Research Foundation) under Germany's Excellence Strategy -- The Berlin 
Mathematics Research Center MATH+ (EXC-2046/1, project ID: 390685689), and by the German Federal Ministry of Education and Research (BMBF) within the Research Campus MODAL (grant numbers 05M14ZAM, 05M20ZBM).}}

\author[1]{Antonia Chmiela\footnote{Corresponding author. Email: 
chmiela@zib.de}$^{,}$}
\author[2]{Elias B. Khalil}
\author[1,3]{Ambros Gleixner}
\author[4]{Andrea Lodi}
\author[1,5]{Sebastian Pokutta}

\affil[1]{Zuse Institute Berlin}
\affil[2]{University of Toronto}
\affil[3]{HTW Berlin}
\affil[4]{Polytechnique Montr\'eal}
\affil[5]{TU Berlin}
\date{}

\usepackage{fancyhdr}
\begin{document}
	
	\maketitle
	
	\begin{abstract}
		\noindent
		Primal heuristics play a crucial role in exact solvers for Mixed 
		Integer Programming (MIP). While solvers are guaranteed to find optimal solutions given sufficient time, real-world applications typically require finding good solutions early on in the search to enable fast decision-making. While much of MIP research focuses on designing effective heuristics, the question of how to manage multiple MIP heuristics in a solver has not received equal attention. Generally, solvers follow hard-coded rules derived from empirical testing on broad sets of instances. Since the performance of heuristics is instance-dependent, using these general rules for a particular problem might not yield the best performance. In this work, we propose the first data-driven framework for scheduling heuristics in an exact MIP 
		solver. By learning from data describing the performance of primal
		heuristics, we obtain a problem-specific schedule of heuristics that 
		collectively find many solutions at minimal cost. We provide a formal 
		description of the problem and propose an efficient algorithm for 
		computing such a schedule. Compared to the default settings of a 
		state-of-the-art academic MIP solver, we are able to reduce the average 
		primal integral by up to $49 \%$ on a class of challenging 
		instances.
	\end{abstract}

	\section{Introduction}
	\label{sec:introduction}
	
	Many decision-making problems arising from real-world applications can be 
	formulated using \textit{Mixed Integer Programming (MIP)}. The 
	\textit{Branch-and-Bound} (\bnb{}) framework is a general approach to solving 
	MIPs 
	to global optimality. Over the recent years, the idea of using  machine learning (ML) to improve optimization techniques has
	gained renewed interest. There exist various approaches to tackle 
	different aspects of the solving process using classical ML techniques. For 
	instance, ML has been used to find good parameter configurations for a 
	solver \cite{hutter09,HutterHoosLeytonbrown2011}, improve node 
	\cite{he14}, variable \cite{khalil16,nair20} or cut 
	\cite{baltean19} selection strategies, and detect decomposable structures 
	\cite{kruber17}.
	
	Even though exact MIP solvers aim for global optimality,
	finding good feasible solutions fast is at least as important, especially 
	in the presence of a time limit. The use of \textit{primal heuristics} is 
	crucial in ensuring good primal performance in modern solvers. For 
	instance, \cite{berthold132} showed that the primal 
	bound--the objective value of the best solution--improved on average by around $80\%$ when primal heuristics were 
	used. Generally, a solver has a variety of primal heuristics implemented, 
	where each class exploits a different idea to find good solutions. 
	During \bnb{}, these heuristics are executed successively
	at each node of the search tree, and improved solutions are reported back to the solver if found. Extensive overviews of different primal heuristics, their computational costs, and their impact in MIP solving can be found in \cite{lodi132,berthold13,berthold18}.

\begin{figure}
		\centering
		\begin{minipage}[t]{.4\textwidth}
		\centering
		\includegraphics[width=1.0\linewidth]{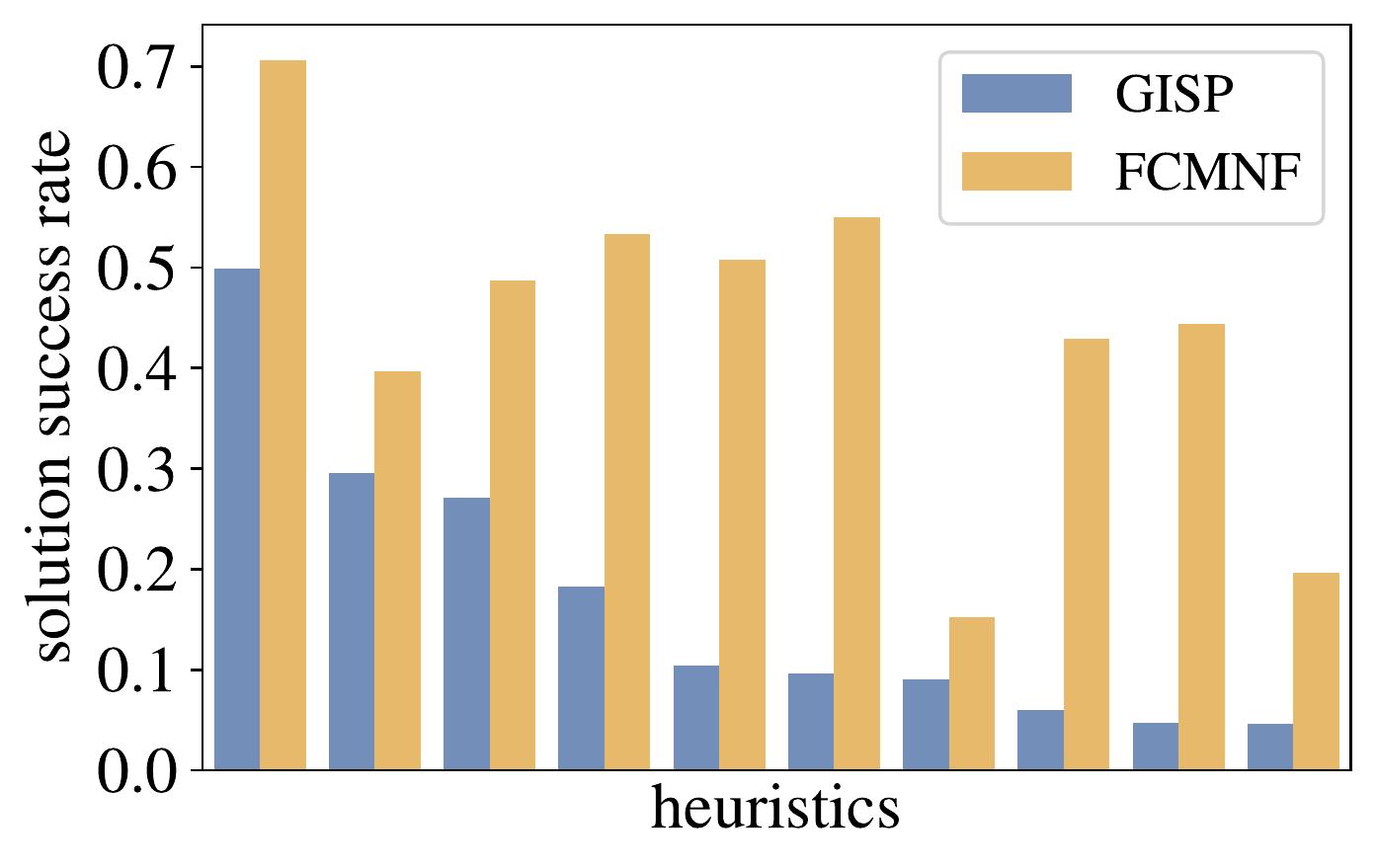}
		\caption{\textbf{Average solution success rates of ten heuristics 
		for two problem classes}. Heuristic success is 
		problem-dependent: each pair of blue-yellow bars belongs to one 
		heuristic, and the heuristics are sorted in descending 
		order w.r.t. the solution success rates for GISP (blue). The yellow 
		bars representing the success rates for FCMNF are far from being 
		sorted, implying that the performance of a heuristic is strongly 
		problem-dependent.}
		\label{fig:heurperformance}
		\end{minipage}%
		\hspace{.05\linewidth}
		\begin{minipage}[t]{.37\textwidth}
		\centering
		\includegraphics[width=1.0\linewidth]{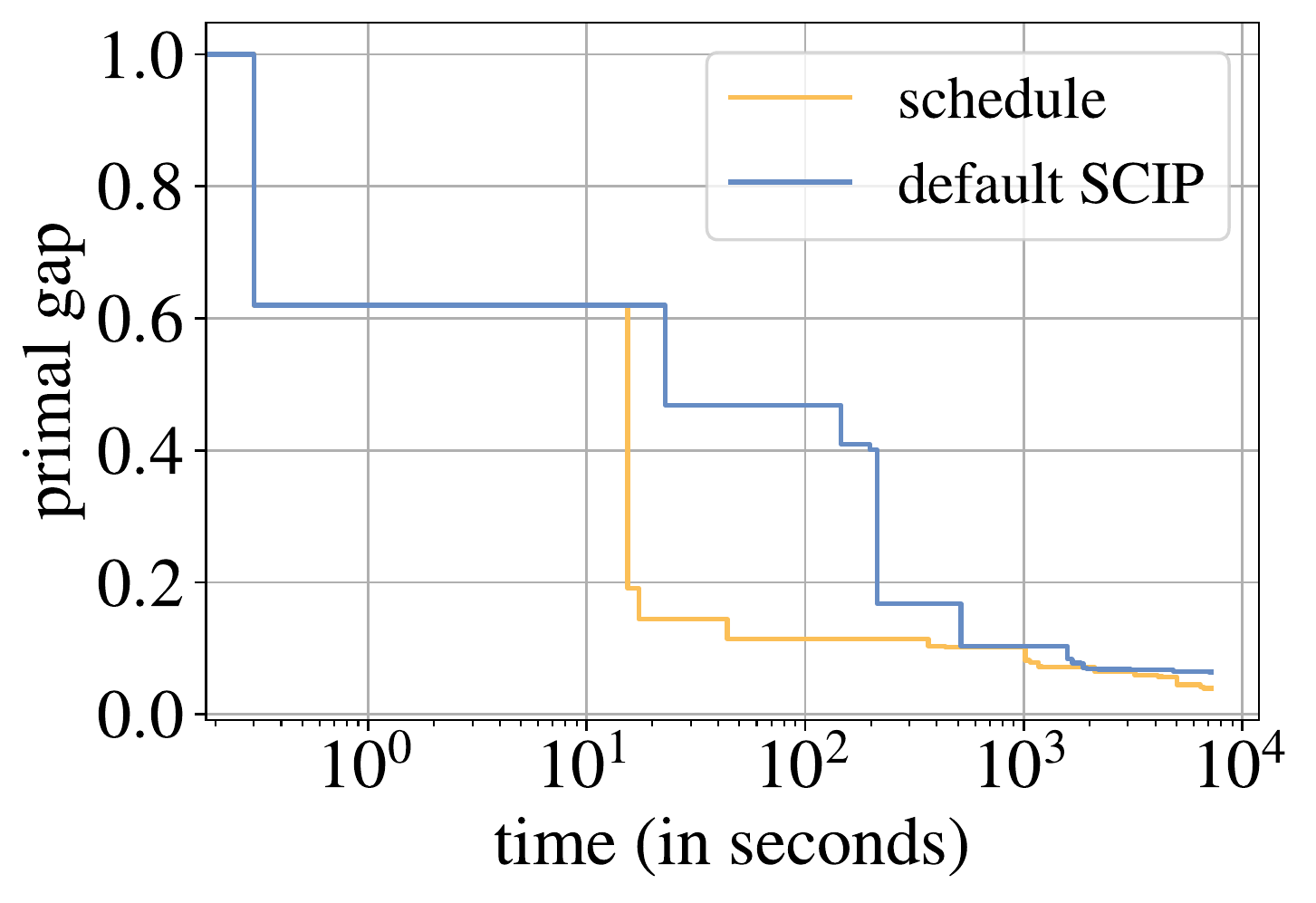}
		\caption{\textbf{Primal gap for an exemplary GISP instance.} Our method's heuristic schedule (orange) obtains better 
		solutions earlier than SCIP's default (blue).}
		\label{fig:primalintegral}
		\end{minipage}
	\end{figure}

   Since most heuristics can be very costly, it is necessary to be strategic about  the order in which the heuristics are executed and the number of iterations allocated to each, with the ultimate goal of obtaining  good primal performance overall.   
	Such decisions are often made by following hard-coded rules derived 
	from testing on broad benchmark test sets. While these static 
	settings yield good performance on average, their performance can be far from 
	optimal when considering specific families of instances.
	To illustrate this fact, Figure \ref{fig:heurperformance} compares the 
	success rates of different primal 
	heuristics for two problem classes:
	the \textit{Generalized Independent Set Problem (GISP)} 
	\cite{hochbaum97,colombi17}
   and the
	\textit{Fixed-Charge Multicommodity Network Flow Problem (FCMNF)} \cite{hewitt10}.
	
	In this paper, we propose a data-driven approach to systematically 
	improve the use of primal heuristics in B\&B. By learning from 
	data about the duration and success of every heuristic call for a set of 
	training instances, we construct a \textit{schedule of 
	heuristics} deciding when and for how long a certain heuristic should be 
	executed to obtain good primal solutions early on. As a result, we are able 
	to significantly improve the use of primal heuristics which is shown 
	in Figure \ref{fig:primalintegral} for one MIP instance.
	
	Even through we will focus on improving primal performance of MIP solving, 
	it is important to note that finding good solutions faster also improves 
	the overall running time of the solver. The B\&B procedure generates a 
	search tree in which the nodes correspond to different subproblems. To 
	determine whether a certain part of the tree 
	should be further explored or pruned, we keep track of the incumbent, i.e., 
	the best feasible solution seen thus far. Hence, when good incumbents are 
	found early on, the size of the search tree may be significantly reduced, 
	leading to the problem being solved faster. On a standardized 
	test set, primal heuristics reduced the solving time by up to $30\%$ 
	\cite{berthold132}.

	\begin{enumerate}
		\item {\textbf{We formalize the learning task} of finding an effective, cost-efficient heuristic schedule on a training dataset as a Mixed Integer Quadratic Program (see Section 
		\ref{sec:formulation});}
		
		\item {We propose an \textbf{efficient heuristic} for solving
			the training (scheduling) problem and a
		\textbf{scalable data collection} strategy (see Section
	\ref{sec:scheduling} and \ref{sec:datacollection});}
		
		\item {We perform \textbf{extensive computational experiments} on a class of challenging instances and \textbf{demonstrate the benefits of our approach} (see 
		Section \ref{sec:expresults}).}
	\end{enumerate}
	Since primal heuristics have such a significant influence on the solving 
	process, optimizing their usage is a topic of ongoing research. For 
	instance, by characterizing nodes with different features, \cite{khalil17} 
	propose an ML method to decide at which nodes heuristics should run to improve 
	primal performance. After that decision, all heuristics are executed 
	according to the predefined rules set by the solver. The authors in 
	\cite{hendel18} and \cite{hendel182} use bandit algorithms to learn from 
	previous calls which heuristics to execute first. In contrast to the method 
	proposed in this paper, their procedure only adapts the order in which 
	heuristics are executed.
	Furthermore, primal performance can also be improved by using 
	hyperparameter tuning \cite{hutter09,HutterHoosLeytonbrown2011}, but 
	generally come with extremely high computational cost, since they do not 
	exploit information about the structure of the problem.
	
	\section{Preliminaries}
	\label{sec:background}
	Let us consider a MIP of the form
	\begin{equation} \label{MIP} \tag{$P_{\textit{MIP}}$}
		\begin{aligned}
			& \underset{x \in \mathbb{R}^n}{\text{min}}
			& & c^\mathsf{T} x, \\
			& \text{s.t.} & &  Ax \leq b, \\
			&   		      & &  x_i \in \mathbb{Z}, \quad \forall i \in I \\
		\end{aligned}
	\end{equation}
	with matrix $A \in \mathbb{R}^{m \times n}$, vectors $c \in \mathbb{R}^n$, 
	$b \in \mathbb{R}^m$, and index set $I \subseteq [n]$.
	A MIP can be solved using Branch-and-Bound, a tree search algorithm that 
	finds an optimal solution to \eqref{MIP} by recursively partitioning the 
	original problem into linear subproblems. The nodes in the resulting search 
	tree correspond to these subproblems. Throughout this work, we assume 
	that each node has a unique index that identifies the node even across branch-and-bound trees obtained for different MIP instances. For a set of instances
	$\mathcal{X}$, we denote the union of the corresponding node indices by 
	$\mathcal{N}_{\mathcal{X}}$.  
	
	\noindent\textbf{Primal Performance Metrics.}
	Since we are interested in finding good solutions fast, we consider a 
	collection of different metrics for primal performance. Beside statistics 
	like the time to the first/best solution and the solution/incumbent success 
	rate, we mainly focus on the \textit{primal integral} \cite{berthold132} 
	as a comprehensive measure of primal performance. Intuitively, this metric can be interpreted as a normalized average of the incumbent value over time. Formally, if $x$ is feasible and $x^*$ is an 
	optimal (or best known)
	solution to
	\eqref{MIP}, the \textit{primal gap} of $x$ is defined as
	\begin{align*}
		\gamma(x) \coloneqq 
		\begin{cases}
			0, &\text{ if } |c^\mathsf{T} x| = |c^\mathsf{T} x^*|, \\
			1, &\text{ if } c^\mathsf{T} x \cdot c^\mathsf{T} x^* < 0, \\
			\dfrac{|c^\mathsf{T} x - c^\mathsf{T} x^*|}{\text{max}\{|c^\mathsf{T} x|,|c^\mathsf{T} x^*|\}}, &\text{ 
				otherwise}.
		\end{cases}
	\end{align*}  
	With $x^t$
	denoting the incumbent at time $t$, the \textit{primal gap 
	function} $p: \mathbb{R}_{\geq 0} \to [0,1]$ is then defined as
	\begin{align*}
		p(t) \coloneqq 
		\begin{cases}
			1, &\text{ if no incumbent is found until time } t, \\
			\gamma(x^t), &\text{ otherwise}.
		\end{cases}
	\end{align*}  
	For a time limit $T \in \mathbb{R}_{\geq 0}$, the primal integral $P(T)$ is then given by the area 
	underneath the primal gap function $p$ up to time $T$,
	\begin{align*}
		P(T) \coloneqq \sum_{i = 1}^{K} p(t_{i - 1})(t_i - t_{i-1}),
	\end{align*}
	where $(K-1)$ incumbents have been found until time $T$, $t_0 = 0$,
	$t_K = T$, and $t_1, \dots, t_{K-1}$ are the points in time at which new incumbents are found.

	Figure~\ref{fig:primalintegral} gives an example for the primal gap function.  The
   primal integrals are the areas under each of the curves. It is easy to see
   that finding near-optimal incumbents earlier shrinks the area under the
   graph of $p$, resulting in a smaller primal integral.

	\section{Data-Driven Heuristic Scheduling}
	\label{sec:formulation}
	
	The performance of heuristics strongly depends on the set of problem 
	instances they are applied to.  Hence, it is natural to consider \textit{data-driven} approaches for optimizing the use of primal heuristics for the instances of interest.
	Concretely, we consider the following practically relevant setting.
   We are given a set of heuristics $\mathcal{H}$ and
	a homogeneous set of training instances $\mathcal{X}$ from the same problem class. In a data collection phase, we are allowed to execute the \bnb{} algorithm on the training instances, observing how each heuristic performs at each node of each search tree. At a high level, our goal is then to leverage this data to obtain a schedule 
	of heuristics that minimizes a primal performance metric.

   The specifics of how such data collection is carried out will be discussed
   later on in the paper.
   First, let us examine the decisions that could potentially benefit from a data-driven approach. Our discussion is inspired by an in-depth analysis of how the source-open academic MIP solver SCIP~\cite{gamrath20} manages primal heuristics. However, our approach is generic and is likely to apply to other MIP solvers.

   \subsection{Controlling the Order}
	\label{sec:ordering}
   One important degree of freedom in scheduling heuristics is the order in which a set of applicable heuristics $\mathcal{H}$ is executed by the solver at a given node.
   This can be controlled by assigning a \textit{priority} for each heuristic.
   In a \textit{heuristic loop}, the solver then iterates over the heuristics in decreasing priority. 
	The loop is terminated if a heuristic finds a new incumbent solution. As such, an ordering $\langle h_{1}, \dots, h_{k} \rangle$ that prioritizes effective heuristics can lead to time savings without sacrificing primal performance.
	
	\subsection{Controlling the Duration}
	\label{subsec:duration}
\begin{figure*}[t]
		\centering
		\includegraphics[width=.415\textwidth]{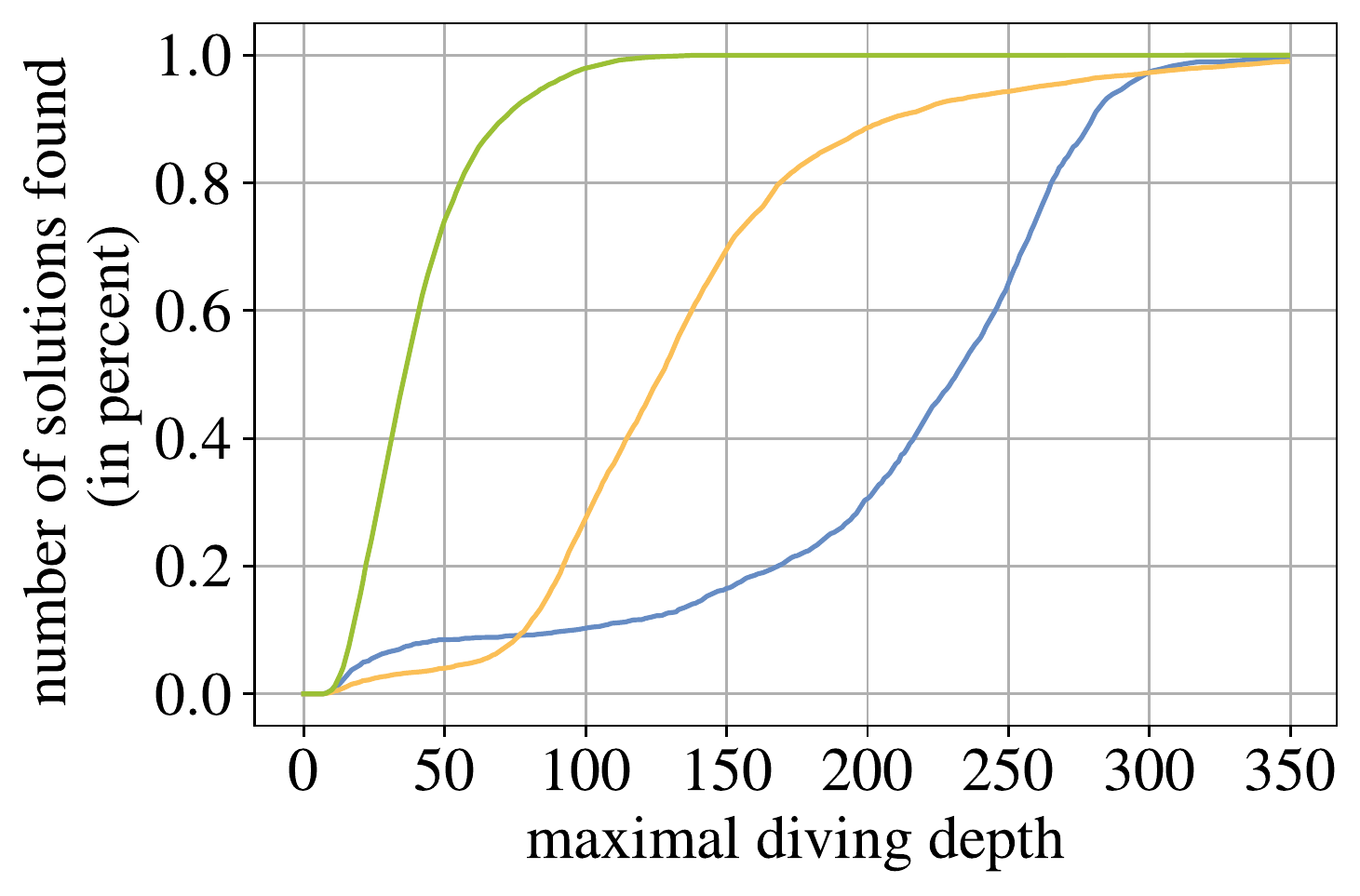}
		\hspace{.05\linewidth}
		\includegraphics[width=.41\textwidth]{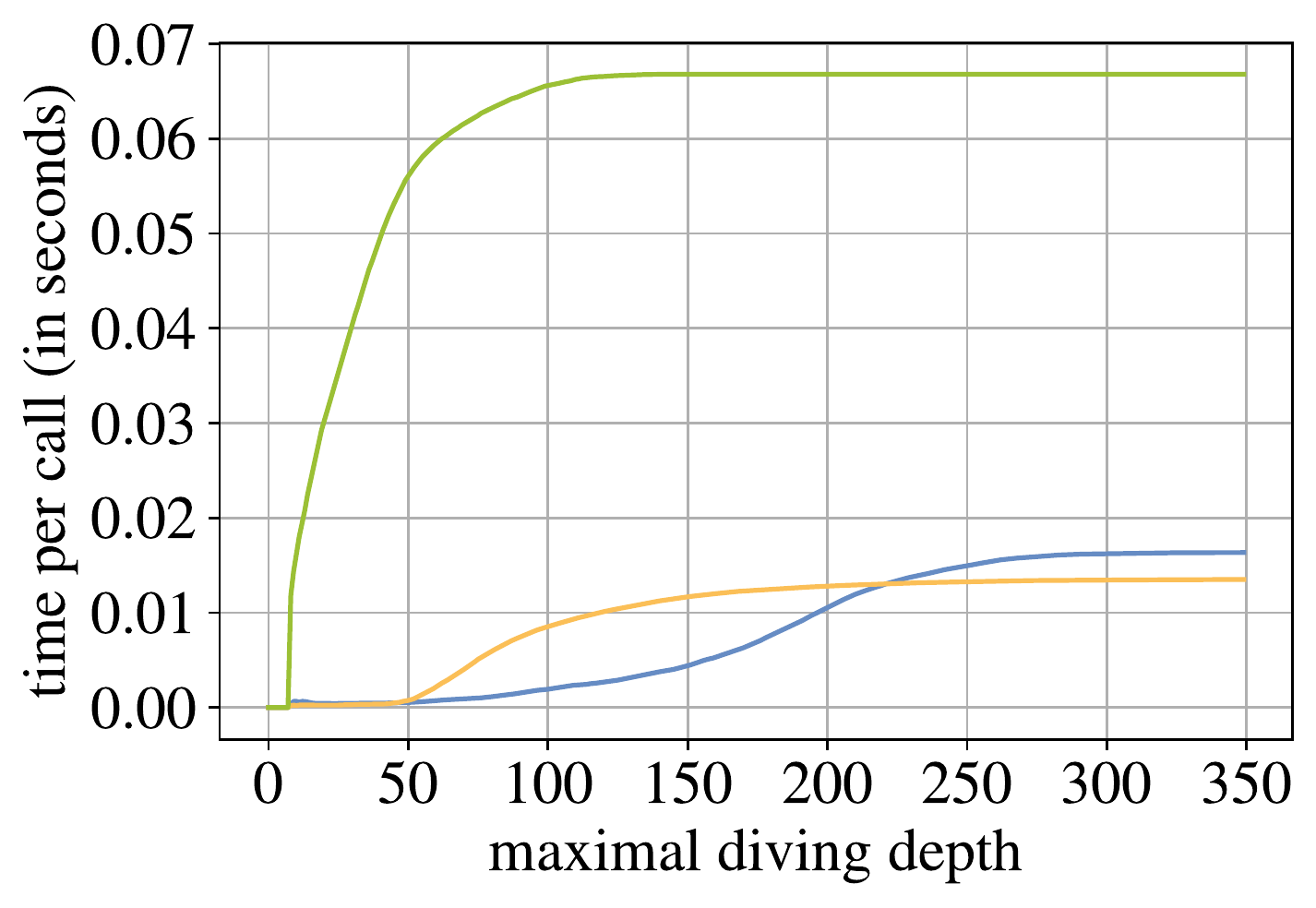}
		\caption{\textbf{Number of solutions found (in percent) and cost of 
		different diving heuristics depending on the the maximal diving depth}: 
		This figure shows the average number of solutions found by a heuristic 
		(left) and average duration in seconds (right) of three diving 
		heuristics when limiting the maximal depth of a dive. Hereby, the 
		baseline for the values on the vertical axis of the left figure is the 
		number of found solutions by the heuristics with no limitations on the 
		diving depth. The 
		likelihood of finding a solution increases with the maximal diving 
		depth. At the same time, an 
		average call to all three heuristics becomes more expensive as the diving depth increases.}
		\label{fig:divingdepth}
	\end{figure*}

	Furthermore, solvers use working limits to control the computational effort spent on heuristics.
	Consider diving heuristics as an example. Increasing the maximal 
	diving depth increases the likelihood of finding an integer feasible
	solution. At the same time, this increases the overall running time. Figure 
	\ref{fig:divingdepth}
	visualizes this cost-benefit trade-off empirically for three different 
	diving heuristics, highlighting the need for a careful ``balancing act".
	For a heuristic $h \in \mathcal{H}$, let $\tau \in 
	\mathbb{R}_{>0}$ denote $h$'s time budget. Then, we are 
	interested in finding a \textit{schedule} $S$ defined by
	\begin{align*}
		S \coloneqq \langle (h_{1}, \tau_{1}), \dots, (h_{k}, \tau_{k}) \rangle, h_i \in \mathcal{H}.
	\end{align*}
	Since controlling the time budget directly can be unreliable and lead to 
	nondeterministic behavior in practice 
	(see Appendix \ref{sec:implementation} for details), a deterministic proxy 
	measure is preferable. 
	For diving heuristics, the maximal diving depth provides a suitable measure 
	as demonstrated by Figure \ref{fig:divingdepth}.
	Similar measures can be used for other types of heuristics, as we will 
	demonstrate with Large Neighborhood Search heuristics in 
	Section~\ref{sec:expresults}. 
   In general, we will refer to $\tau_i$ as the maximal number of 
	\emph{iterations} that is alloted to a heuristic~$h_i$ in schedule~$S$.
	
	\subsection{Deriving the Scheduling Problem}
	Having argued for order and duration as suitable control decisions, we will now formalize our heuristic scheduling problem. 
	Ideally, we would like to construct a single schedule $S$ that minimizes the 
	primal integral, as defined in Section \ref{sec:background}, averaged over the training instances $\mathcal{X}$. Unfortunately, it is very difficult to optimize the primal integral directly, as it depends on the \textit{sequence} of incumbents found over time during \bnb{}. The primal integral also depends on the way the search tree is explored, which is affected by pruning, further complicating any attempt at directly optimizing this primal metric. 
	
	We address this difficulty by considering a more practical surrogate objective. Recall that 
	$\mathcal{N}_{\mathcal{X}}$ denotes the collection of search tree nodes of 
	the set of training instances $\mathcal{X}$. We will construct a schedule $S$ that 
	finds feasible solutions for a large fraction of the nodes in 
	$\mathcal{N}_{\mathcal{X}}$, while also minimizing the number of iterations 
	spent by schedule $S$. Note that we consider feasible solutions instead of 
	incumbents here: This way, we are able to obtain more data faster since a 
	heuristic finds a feasible solution more often than a new incumbent. The 
	framework we propose in the following can handle incumbents instead, but we 
	have found no benefit in that in preliminary experiments.

	For a heuristic $h \in \mathcal{H}$ and node $N$, denote by $t(h,N)$
	the iterations necessary for $h$ to find a
	solution at node $N$, and set $t(h,N) = \infty$ if $h$ does not succeed at $N$.
   Now suppose a schedule $S$ is successful at node $N$, i.e., some heuristic finds a solution within the budget allocated to it in $S$. Let
   \[
   j_S = \text{min}\{j = 1,\ldots,|\mathcal{H}| \mid t(h_j,N) \leq \tau_j \}
   \]
   be the index of the first successful heuristic.  Following the (successful) execution of $h_{j_S}$, the
   heuristic loop is terminated, and the time spent by schedule $S$ at node
   $N$ is given by
	\begin{align*}
	  T(S,N) \coloneqq \sum_{i = 1}^{j_S-1} \tau_i + t(h_{j_S},N).
	\end{align*}
   Otherwise, set $T(S,N) \coloneqq \sum_{i = 1}^{k} \tau_i + 1$\footnote{We 
   add $1$ to penalize unsolved nodes.}.

   Furthermore, let $\mathcal{N}_S$ denote the set of nodes at which schedule
   $S$ is successful in finding a solution.
	Then, we consider the heuristic scheduling problem given by
	\begin{equation} \label{eq:schedulingproblem} \tag{$P_{\mathcal{S}}$}
			\underset{S \in \mathcal{S}}{\text{min}}
			\sum_{N \in \mathcal{N}_{\mathcal{X}}} T(S,N)
			\;\text{ s.t. }\;
			|\mathcal{N}_S| \geq \alpha |\mathcal{N}_{\mathcal{X}}|.
	\end{equation}
	Here $\alpha \in [0,1]$ denotes a minimum fraction of nodes at which we want the schedule to find a solution.
	Problem \eqref{eq:schedulingproblem} can be formulated as a Mixed-Integer 
	Quadratic Program (MIQP); the exact formulation can be found in 
	Appendix \ref{sec:mip}.
	
	To find such a schedule, we need to know $t(h,N)$ for every heuristic $h$ 
	and node $N$. Hence, when collecting data for the instances in the training 
	set $\mathcal{X}$, we track for every B\&B node $N$ at which a 
	heuristic $h$ was called, the number of iterations $\tau_N^h$ it took $h$ to 
	find a feasible solution; we set $\tau_N^h = \infty$ if $h$ does not succeed at $N$. 
	Formally, we require a training dataset
	\begin{align*}
		\mathcal{D} \coloneqq \{ (h, N, \tau_N^h)) \mid h \in \mathcal{H}, N \in 
		\mathcal{N}_{\mathcal{X}}, \tau_N^h \in \mathbb{R}_+ \cup \{ \infty \} 
		\}.
	\end{align*}
	Section \ref{sec:datacollection} describes a computationally efficient 
	approach for building $\mathcal{D}$ using a \textit{single} \bnb{} run per 
	training instance.  

	\section{Solving the Scheduling Problem}
	\label{sec:scheduling}
	
    Problem \eqref{eq:schedulingproblem} is a generalization of the Pipelined 
    Set Cover Problem which is known to be $\mathcal{NP}$-hard 
    \cite{munagala05}. As for the MIQP in Appendix \ref{sec:mip}, tackling it 
    using a 
    non-linear integer programming solver is challenging: the MIQP has 
    $O(|\mathcal{H}||\mathcal{N}_{\mathcal{X}}|)$ variables and constraints, 
    and a single training instance may involve thousands of search tree nodes, 
    leading to an MIQP with hundreds of thousands of variables and constraints even 
    with a handful of heuristics and tens of training instances.
    
    As already mentioned in the beginning, one approach to finding a schedule 
    that heuristically solves \eqref{eq:schedulingproblem} is using a 
    hyperparameter tuning software like SMAC \cite{HutterHoosLeytonbrown2011}. Since SMAC is a 
    sequential algorithm that searches for a good parameter configuration by 
    successively adapting and re-testing its best settings, training a SMAC 
    schedule can get very expensive quickly. In the following, we present a more 
    efficient approach. 
    
    We now direct our attention towards designing an efficient heuristic 
    algorithm for~\eqref{eq:schedulingproblem}. 
	A similar problem was studied by 
	\cite{streeter07} in the context of decision problems. Among other things, 
	the author discusses how 
	to find a schedule of (randomized) heuristics that minimizes the expected 
	time necessary to solve a set of training instances $\mathcal{X}$ of a decision problem. Although this setting is somewhat similar to ours, there exist multiple aspects in which they differ 
	significantly:
	\begin{enumerate}
		\item \textit{Decision problems are considered instead of MIPs:} 
		Solving a MIP is generally much more challenging than solving a 
		decision problem.
		When solving a MIP with B\&B, we normally have to solve 
		many linear subproblems. Since in theory, every such LP is an 
		opportunity for a heuristic to find a new incumbent, we consider the 
		set of nodes $\mathcal{N}_{\mathcal{X}}$ instead of $\mathcal{X}$ as 
		the ``instances'' we want to solve.
		
		\item \textit{A heuristic call can be suspended and resumed:} In the 
		work of~\cite{streeter07}, a heuristic can be executed in a 
		``suspend-and-resume 
		model'': If $h$ was executed before, the action $(h, \tau)$ represents 
		\textit{continuing} a heuristic run for an additional $\tau$ iterations. 
		When $h$ reaches the iteration limit, the run is suspended and its state kept in memory such that it can be resumed later in the schedule. 
		The ``suspend-and-resume" model is not used in MIP solving due to challenges in maintaining the states of heuristics in memory. As such, we allow every heuristic to be included in the schedule at most once.
		\item \textit{Time is used to control the duration of a heuristic run:}
        Controlling time directly is unreliable in practice and can lead to
        nondeterministic behavior of the solver.  Instead, we rely on different
        proxy measures for different classes of
		  heuristics. Thus, when building a schedule that contains heuristics of 
		distinct types (e.g., diving and LNS heuristics), we need to ensure that these measures are comparable. 
	\end{enumerate}
	Despite these differences, it is useful to examine the greedy scheduling 
	approach proposed by~\cite{streeter07}. A
	schedule is built by successively adding the action $(h,\tau)$ to $G$ 
	that maximizes the ratio of the marginal increase in the number of instances solved to the cost (i.e., $\tau$) of 
	including $(h, \tau)$. As shown in Corollary 2 of \cite{streeter07}, the 
	\textit{greedy schedule} $G$ yields a 4-approximation of that version of the 
	scheduling problem. In an attempt to leverage this elegant heuristic in our problem~\eqref{eq:schedulingproblem}, we will describe it formally.
	
	Let us denote the greedy schedule by $G \coloneqq \langle g_1, \dots, g_k 
	\rangle$. Then, $G$ is defined inductively by setting $G_0 = \langle 
	\rangle$ and $G_j = \langle g_1, \dots, g_{j} \rangle $ with
	\begin{equation*}
		\begin{aligned}
			g_j = \underset{(h,\tau) \in \mathcal{H}_{j-1} \times 
			\mathcal{T}}{\text{argmax}} \frac{|\{ N \in 
			\mathcal{N}_{\mathcal{X}}^{j-1} \mid \tau_N^h \leq \tau\}|}{\tau}.
		\end{aligned}
	\end{equation*}
	Here, $\mathcal{H}_i$ denotes the set of heuristics that are not yet in 
	$G_i$,
	$\mathcal{N}_{\mathcal{X}}^{i}$ denotes the subset of nodes where $G_i$ is not yet successful in finding a solution, and 
	$\mathcal{T}$ is the interval generated by all possible iteration limits in 
	$\mathcal{D}$, i.e.,
		\begin{align*}
			\mathcal{T} \coloneqq [\text{min}\{\tau_N^h \mid (N, h, \tau_N^h) \in 
			\mathcal{D}\}, \text{max}\{\tau_N^h \mid (N, h, \tau_N^h) \in 
			\mathcal{D}\}].
		\end{align*}
	We stop adding actions $g_j$ when $G_j$ finds a solution at all nodes
	in $\mathcal{N}_{\mathcal{X}}$ or all heuristics are 
	contained in the schedule, i.e., $\mathcal{H}_j = \emptyset$. 
	
	Unfortunately, we can show that the resulting schedule can perform arbitrarily bad in our setting. Consider the following situation. We assume that there are 
	100 nodes in $\mathcal{N}_{\mathcal{X}}$ and only one heuristic $h$. This 
	heuristic solves one node in just one 
	iteration and takes 100 iterations each for the other 99 nodes. Following 
	the greedy approach, the resulting schedule would be $G = \langle (h,1) 
	\rangle$ since $\frac{1}{1} > \frac{99}{100}$. Whenever $\alpha > 0.01$, 
	$G$ would be infeasible for our constrained problem~\eqref{eq:schedulingproblem}. Since we are not allowed to add a heuristic more 
	than once, this cannot be fixed with the current algorithm.
	
	To avoid this situation, we propose the following modification. Instead 
	of only considering the heuristics that are not in $G_{j-1}$ when choosing 
	the next action $g_j$, we also consider the option to run the last 
	heuristic $h_{j-1}$ of $G_{j-1}$ for longer. That is, we allow to choose 
	$(h_{j-1}, \tau)$ with $\tau > \tau_{j-1}$. Note that the cost of 
	adding $(h_{j-1}, \tau)$ to the schedule is not $\tau$, but $\tau - 
	\tau_{j-1}$, since we decide to run $h_{j-1}$ for $\tau - \tau_{j-1}$ 
	iterations longer and not to rerun $h_{j-1}$ for $\tau$ iterations.
	
\begin{figure}[t]
		\centering
		\includegraphics[width=.41\textwidth]{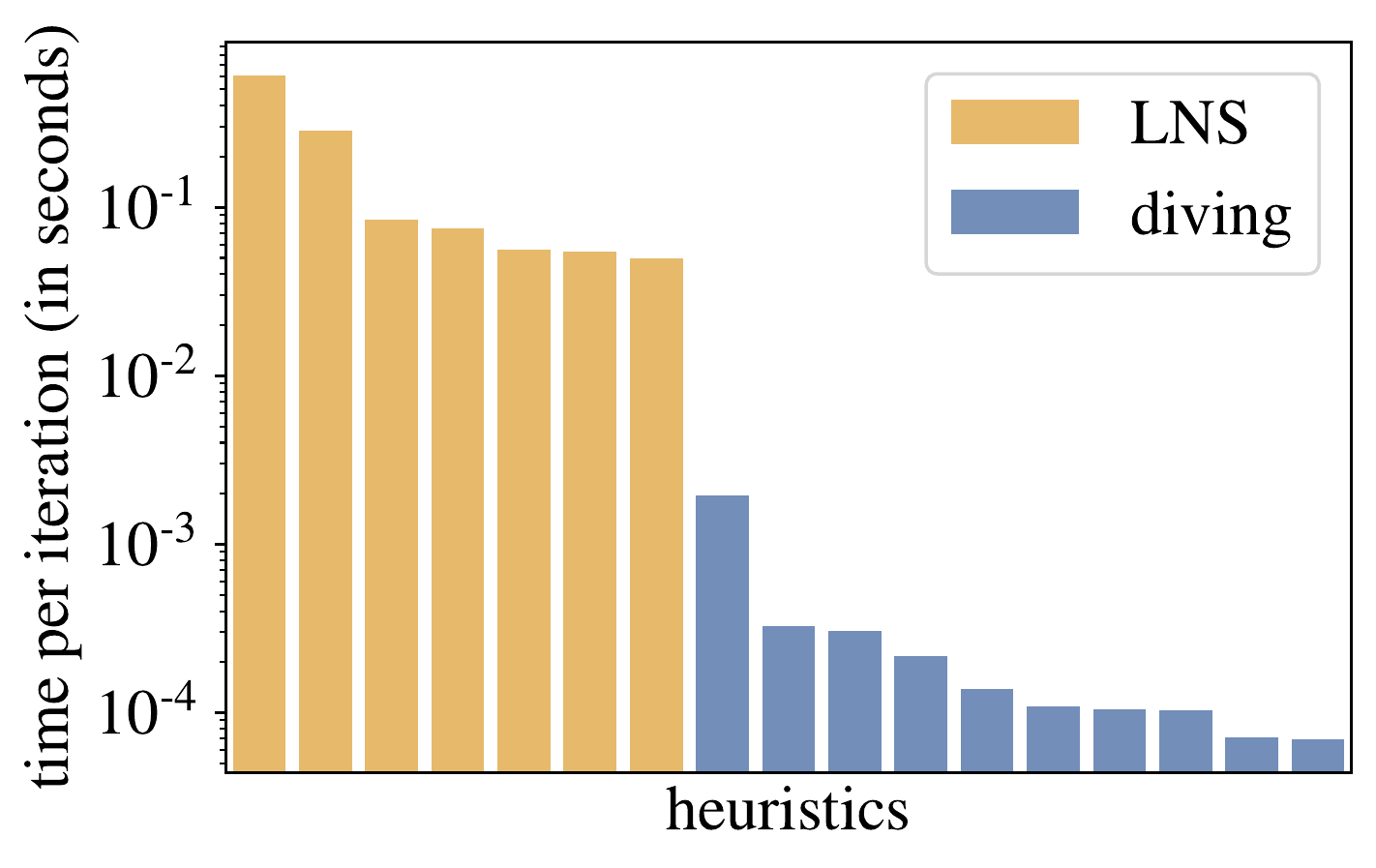}
		\caption{\textbf{Comparison of average cost of iterations for 
				different primal heuristics:} While the cost of an iteration is 
			relatively similar among heuristics of the same type, they differ 
			significantly when comparing diving and LNS with each other. On 
			average, an iteration for LNS heuristics (number of nodes in 
			sub-MIP) 
			is much more expensive than for diving heuristics (maximal diving 
			depth).}
		\label{fig:heurcost}
	\end{figure}

	Furthermore, when including different classes of heuristics in the 
	schedule, the respective time measures are not necessarily comparable (see 
	Figure \ref{fig:heurcost}). To circumvent this problem, we use the average 
	time per iteration to normalize different notions of iterations. In the 
	following, we denote the average cost of an iteration by $t^h_{avg}$ for 
	heuristic $h$. Note that $t^h_{avg}$ can be easily computed by also 
	tracking the duration (w.r.t. time) of a heuristic run in data collection.
	Hence, we redefine $g_j$ and obtain
	\begin{equation*}
		\begin{aligned}
			g_j = \underset{(h,\tau) \in \mathcal{A}_{j-1}}{\text{argmax}} 
			\frac{|\{ 
			N \in \mathcal{N}_{\mathcal{X}}^{j-1} \mid \tau_N^h \leq 
			\tau\}|}{c_{j-1}(h,\tau)},
		\end{aligned}
	\end{equation*}
	with $\mathcal{A}_i \coloneqq (\mathcal{H}_i \times \mathcal{T}) \cup \{ 
	(h_i, \tau) \mid \tau > \tau_i, \tau \in \mathcal{T}\}$ and
	\begin{align*}
		c_{i}(h,\tau) \coloneqq
		\begin{cases}
			t^h_{avg} \tau, &\text{ if $h \neq h_{i}$} \\
			t^h_{avg} (\tau - \tau_{i}), &\text{ otherwise}.
		\end{cases}
	\end{align*}
	We set $\mathcal{A}_0 \coloneqq \mathcal{H} \times \mathcal{T}$ and 
	$c_0(h, \tau) = t^h_{avg} \tau$. With this modification, we would obtain 
	the schedule 
	$G = \langle (h,100) \rangle$ (which solves all 100 nodes) in the above 
	example.

   Finally, note that this greedy procedure still does not explicitly enforce that the 
	schedule is successful at a fraction of at least $\alpha$ nodes. In our experiments, 
	however, we observe that the resultings schedules reach a success rate of 
	$\alpha=98\%$ or above. The final formulation of the 
	algorithm can be found in Algorithm \ref{alg:greedy}.
	
\begin{algorithm}[t]
	\caption{Greedy algorithm to obtain a schedule}
	\label{alg:greedy}
	\begin{algorithmic}
		\STATE {\bfseries Input:} Nodes $\mathcal{N}_\mathcal{X}$, 
		heuristics $\mathcal{H}$, data $D$, time frame $\mathcal{T}$
		\STATE {\bfseries Output:} Greedy Schedule $G$
		
		\STATE $G \gets \langle \rangle$
		\STATE $\mathcal{N}_{unsol} \gets \mathcal{N}_\mathcal{X}$
		\STATE $improve \gets$ TRUE
		
		\REPEAT
		
		\STATE $(h^*,\tau^*) \gets \underset{(h,\tau) \in 
			\mathcal{A}}{\text{argmax}} \left[ \frac{ | \{ N 
			\in 						\mathcal{N}_{unsol} \mid \tau_h^N \leq 
			\tau \} |}{c(h,\tau)} \right]$
		
		\IF{$\frac{ | \{ N \in \mathcal{N}_{unsol} \mid \tau_{h^*}^N \leq 
				\tau^* \} |}{c(h,\tau^*)} > 0$}
		\STATE $G \gets G \oplus \langle(h^*,\tau^*) \rangle$
		\STATE $\mathcal{N}_{unsol} \gets \mathcal{N}_{unsol} \setminus \{ 
		N \in \mathcal{N}_{unsol} \mid \tau_{h^*}^N \leq \tau^* \}$
		\ELSE
		\STATE $improve \gets$ FALSE
		\ENDIF
		
		\UNTIL{$improve ==$ FALSE}
	\end{algorithmic}
\end{algorithm}
	
\begin{figure}[H]
		\centering
		\includegraphics[width=.6\textwidth]{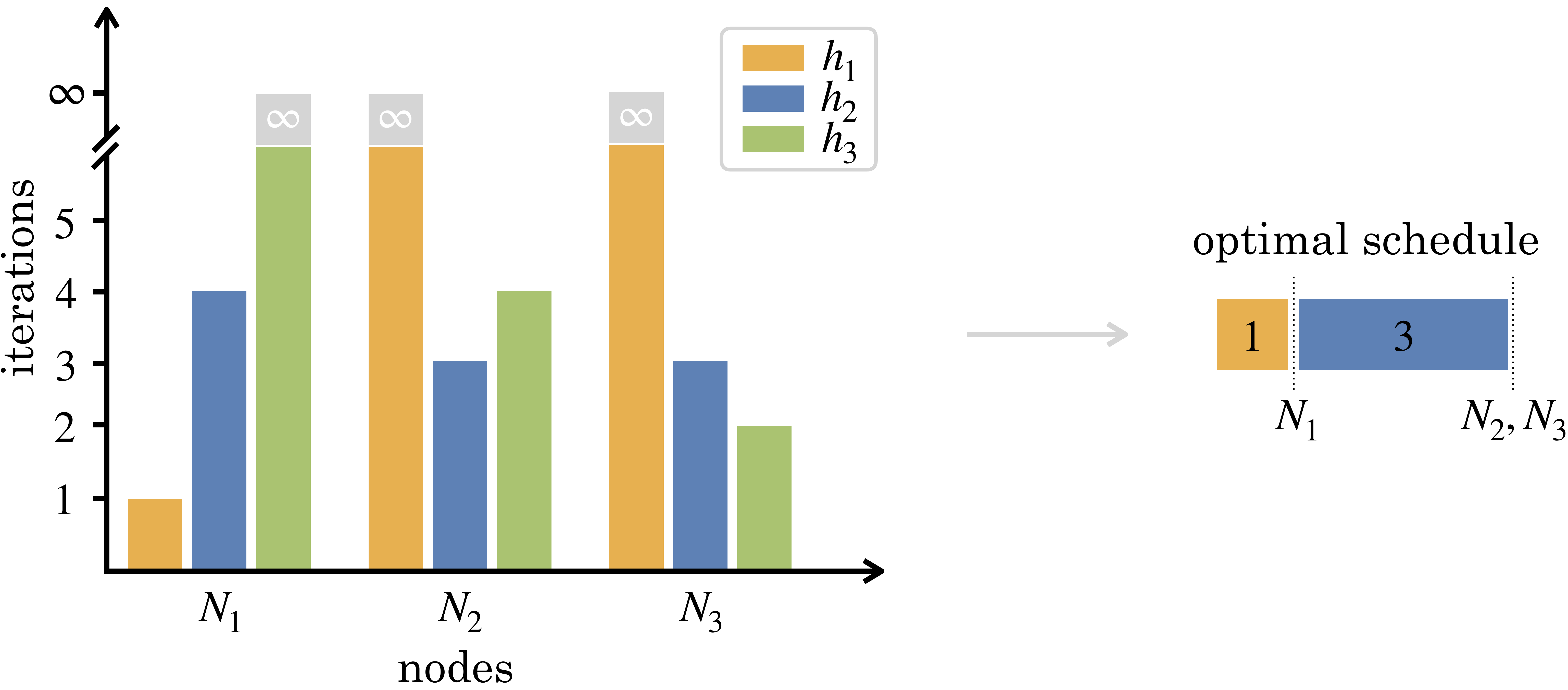}
		\caption{\textbf{Example of how to obtain a heuristic schedule from 
		data:} The data is shown on the left for three heuristics and nodes and 
		the (optimal) schedule obtained by following Algorithm \ref{alg:greedy} 
		is 
		illustrated on the right.}
		\label{fig:example}
	\end{figure}

	\textbf{Example.} Figure \ref{fig:example} shows an example of how we 
	obtain a schedule with three heurisitcs and nodes. As the left figure 
	indicates, the data set is given by
	\begin{align*}
		\mathcal{D} = \{&(h_1, N_1, 1), (h_1, N_2, \infty), (h_1, N_3, 
		\infty), (h_2, N_1, 4), (h_2, N_2, 3), \\
		&(h_2, N_3, 3), (h_3, N_1, 
		\infty), (h_3, N_2, 4), (h_3, N_3, 2) \}.
	\end{align*}
	Let us now assume that an iterations of each heuristic has the same 
	average costs, i.e., $t^{h_1}_{avg} = t^{h_2}_{avg} = t^{h_3}_{avg}$, we 
	build an schedule $G$ as follows. First, we add the action $(h_1,1)$, since 
	$h_1$ solves one node with only one iteration yielding a ratio that cannot 
	be bet by the other heuristics. No other node can be solved by $h_1$, hence 
	it does not have to be considered anymore, as well as node $N_1$. Among the 
	remaining possibilities, the action $(h_2, 3)$ is the best, since $h_2$ 
	solves both nodes in three iterations yielding a ratio of $\frac{2}{3}$. In 
	contrast, executing $h_3$ for two and four iterations, respectively, would 
	yield a ratio of $\frac{1}{2}$. Since this is smaller, we add $(h_2, 3)$ to 
	the $G$ and obtain the schedule $G = \langle (h_1,1), (h_2,3) \rangle$ 
	which solves all three nodes as shown on the right of Figure 
	\ref{fig:example}. It is easy to see that this schedule is an optimal 
	solution of \eqref{eq:schedulingproblem} for $\alpha > \frac{1}{3}$.
	
	\section{Data Collection}
	\label{sec:datacollection}
    The scheduling approach described thus far rests on the availability of a 
    training data set of the form
	\begin{align*}
		\mathcal{D} \coloneqq \{ (h, N, \tau_N^h) \mid h \in \mathcal{H}, N \in 
		\mathcal{N}_{\mathcal{X}}, \tau_N^h \in \mathbb{R}_+ \cup \{ \infty \} 
		\}.
	\end{align*}
    In words, each entry in data set $\mathcal{D}$ is a triplet containing: the 
    index of a heuristic $h$; the index of a \bnb{} node $N$ coming from one of 
    the training instances in $\mathcal{X}$; the number of iterations required 
    by $h$ to find a feasible solution at the node $N$. The latter piece of 
    information, $\tau_N^h$, must be collected by executing the heuristic and 
    observing its performance. Two main challenges arise in collecting such a 
    data set for multiple heuristics:
	\begin{enumerate}
		\item \textit{Efficient data collection:} 
        Solving $\mathcal{NP}$-hard MIPs by \bnb{} remains
        computationally expensive, even given the sophisticated techniques
        implemented in today's solvers.  This poses difficulties to ML
        approaches that create a single reward signal from one MIP evaluation,
        which may take several minutes up to hours.  This holds in particular
        for challenging problem classes that are the focus of this work. In 
        other words, even with a handful of heuristics, i.e., a small set 
        $\mathcal{H}$, it is prohibitive to run \bnb{} once for each 
        heuristic-training instance pair in order to construct the data set 
        $\mathcal{D}$.

		\item \textit{Obtaining unbiased data:}
		On the other hand, executing multiple heuristics at each node of the search tree during data collection can have dangerous side effects: if a heuristic finds an incumbent, subsequent heuristics are no longer executed at the same node, as described in Section~\ref{sec:ordering}.
	\end{enumerate}
   We address the first point by using a specially crafted version of the MIP
   solver for collecting \emph{multiple reward signals} for the execution of
   \emph{multiple heuristics} per single MIP evaluation during the training
   phase.  As a result, we obtain a large amount of data points that scales with
   the running time of the MIP solves.  This has the clear advantage that the
   efficiency of our data collection does not automatically decrease when the
   time to evaluate a single MIP increases for more challenging problems.

   To address the second point and prevent bias from mutual interaction of
   different heuristic calls during training, we engineered the MIP solver to be
   executed in a special \emph{shadow mode}, where heuristics are called in a
   sandbox environment and interaction with the main solving path is maximally
   reduced.  In particular this means that new incumbents and primal bounds are
   not communicated back, but only recorded for training data. This setting is 
   an improved version of the shadow mode introduced in \cite{khalil17}.
   
   As a result of these measures, we have instrumented the SCIP solver in a way 
   that allows for the collection of a proper data set $\mathcal{D}$ with a 
   \textit{single run} of the Branch-and-Bound algorithm per training instance.
	
	\section{Computational Results}
	\label{sec:expresults}
	We will now detail our computational results. The code we used for 
	data collection and scheduling is publicly available 
	\footnote{\url{https://github.com/antoniach/heuristic-scheduling}}.
	
	\subsection{Heuristics}
	\label{sec:heuristics}
	We can build a schedule containing arbitrary heuristics as long as they 
	have some type of time measure. In this work, we focus on two 
	broad groups of heuristics: \textit{Diving} and \textit{Large Neighborhood 
	Search (LNS)}. Both classes are much more computationally expensive than 
	simpler heuristics like rounding, but are generally also more likely to 
	find (good) solutions \cite{berthold06}. That is why it is particularly 
	important to schedule 
	these heuristics most economically.

	\textbf{Diving Heuristics.} Diving heuristics examine 
	a single probing path by successively fixing variables according to a 
	specific rule.
	There are multiple ways of controlling the duration of a dive. After 
	careful consideration of the different options, we decided on using the 
	maximum diving depth to limit the cost of a call to a diving heuristic: It 
	is both related to the effort spent by the heuristic and its likelihood of 
	success.
	
	\textbf{LNS Heuristics.} This class of heuristics first builds a 
	neighborhood of some reference point
	which is then searched for improving solutions by solving a sub-MIP. 
	To control the duration of a call to a LNS heuristic, we choose to limit 
	the number of nodes in the sub-MIP.
	The idea behind this measure is similar to limiting the diving depth of 
	diving heuristics: In both cases, we control the number of subproblems that 
	a heuristic considers within its execution. Nevertheless, the two measures are 
	not directly comparable, as shown in Figure \ref{fig:heurcost}. 
	
	To summarize, we use 16 primal heuristics in our schedule: ten diving and 
	six LNS heuristics. 
	By controlling this set, we cover the majority of the
	more complex heuristics implemented in SCIP. 
	All other heuristics are executed after the schedule according to their 
	default settings.
	
	\subsection{Instances}
	\label{sec:instances}
	Since our goal is to improve the primal performance of a solver, we focus 
	on a primally challenging problem class: The 
	\textit{Generalized Independent Set Problem (GISP)} 
	\cite{hochbaum97,colombi17}.
	In the following, we will briefly explain how we generate and partition the 
	instances.
	
	Let $G = (V,E)$ be a graph and $E' \subseteq E$ a subset of 
	removable edges. Each vertex has a revenue and every edge has
	a cost associated with it. Then, GISP asks to select a subset of 
	vertices and removable edges that maximizes the profit, i.e., the 
	difference of vertex revenue and edge costs. Thereby, no edge should exist 
	between two selected vertices $v,u \in V$, i.e., either we have that $(v,u) 
	\notin E$ or $(v,u) \in E'$ is removed.    
	
	We generate
	GISP instances in the following way. Given a graph, we 
	randomize the set of removable edges by setting the probability that an 
	edge is in $E'$ to $\alpha = 0.75$. Furthermore, we choose the revenue for 
	each node to be $100$ and the cost of every edge as $1$. This results in a 
	configuration for which it is difficult to find good feasible solutions as 
	shown in \cite{colombi17}.
	
	We use this scheme to generate two types of instances. The first one takes 
	graphs from the 1993 DIMACS Challenge which is also used by \cite{khalil17, 
	colombi17}. Thereby, we focus on the same twelve dense graphs as well as 
	the same train/test partition as in \cite{khalil17}.
	The training set consists of six graphs with 125--300 nodes and 
	6963--20864 edges, whereas the testing graphs are considerably larger with 
	250--400 nodes and 21928--71819 edges. We generate 20 instances for every 
	graph by using different random seeds, 
	leaving us with 120 instances for training as well as testing.
	For the second group of GISP instances, we use randomly generated graphs 
	where the number of nodes is uniformly chosen from $\{L, \dots, U\}$ for 
	bounds $L,U \in \mathbb{N}$. An edge is added to the resulting graph with 
	probability $\bar{\alpha} = 0.1$, giving us slightly less dense graphs than 
	the previous case.
	We denote these sets by \textsc{[L,U]}. For each set, we generate 25 
	instances for training and 10 for testing. The smallest set of graphs then 
	has 150--160 nodes and 1099--1268 edges whereas the largest set consists of 
	graphs with 550--560 nodes and 14932--15660 edges.

	\subsection{Results}
	\label{sec:results}
	To study the performance of our approach, we used the state-of-the-art 
	solver SCIP 7.0 \cite{gamrath20} with CPLEX 12.10.0.0 
	\footnote{\url{https://www.ibm.com/products/ilog-cplex-optimization-studio}}
	as the underlying LP solver. Thereby, we needed to modify 
	SCIP's source code to 
	collect data as described in Section \ref{sec:datacollection}, as well as 
	control heuristic parameters that are not already implemented by default. 
	For our experiments, we used a Linux cluster of Intel Xeon CPU E5-2660 v3 
	2.60GHz with 25MB cache and 128GB main memory. The time limit in all 
	experiments was set to two hours; for data collection, we used a time limit 
	of four hours. Since the primal integral depends on time, we
	ran one process at a time on every machine, allowing for accurate on time measurements.
	
    MIP solver performance can be highly sensitive to even small and seemingly 
    performance-neutral perturbations during the solving process~\cite{lodi13}, 
    a phenomenon referred to as \textit{performance variability}. We 
    implemented a more exhaustive testing framework than
 	the commonly used benchmark methodology in MIP that uses extensive 
    cross-validation in addition to multiple random seeds.
    
    In addition to comparing our scheduling method against default SCIP, we 
    also compare against \textsc{scip\_tuned}, a hand-tuned version of SCIP's 
    default settings for GISP\footnote{We set the frequency offset to 0 for all 
    diving heuristics.}. Since in practice, a MIP expert would try to 
    manually optimize some parameters when dealing with a homogeneous set of 
    instances, we emulated that process to create an even stronger baseline to 
    compare our method against.
 
	\textbf{Random graph instances.} Table \ref{table:crossvalidation} shows 
	the results of the cross-validation experiments for schedules with 
	diving heuristics. Our scheduling framework yields a significant improvement w.r.t. primal 
	integral on all test sets. Since this improvement is consistent over all 
	schedules and test sets, we are able to validate that the behavior 	
	actually comes from our procedure. Especially remarkable is the fact that 
	the schedules trained on smaller instances also perform well on much larger 
	instances.

	Note that the instances in the first three test 
	sets were solved to optimality by all settings whereas the remaining ones 
	terminated after two hours without a provably optimal solution. When 
	looking at the instances that were not solved to optimality, we can see 
	that the schedules perform especially well on instances of increasing 
	difficulty. This behavior is intuitive: Since our method aims to improve 
	the primal performance of a solver, it performs better when an instance is very
	challenging on the primal side.
	
	\begin{table}
		\renewcommand{\arraystretch}{1.35}
		\tiny
		\centering
		
		\begin{tabular}{lcccccccccc}
			\hline
			\hline
			\diagbox[width=6em]{train}{test} & \textsc{[150,160]} & 
			\textsc{[200,210]} & 
			\textsc{[250,260]} & \textsc{[300,310]} & \textsc{[350,360]} & 
			\textsc{[400,410]} & \textsc{[450,460]} & \textsc{[500,510]} & 
			\textsc{[550,560]} \\
			\hline
			\textsc{[150,160]} & $0.89 \pm 0.23$ &  $0.76 \pm 0.22$ &  $0.87 
			\pm 0.37$ &  
			$0.95 \pm 0.40$ & $0.87 \pm 0.28$ &  $0.86 \pm 0.23$ &  $0.78 \pm 
			0.24$ &  
			$0.80 \pm 0.25$ &  
			$0.65 \pm 0.24$ \\
			\textsc{[200,210]} & $0.94 \pm 0.28$ &  $0.75 \pm 0.25$ &  $0.82 
			\pm 0.30$ &  
			$0.91 \pm 0.34$ & $0.93 \pm 0.23$ &  $0.90 \pm 0.28$ &  $0.83 \pm 
			0.22$ &  
			$0.79 \pm 0.20$ &  
			$0.66 \pm 0.20$ \\
			\textsc{[250,260]} & $0.89 \pm 0.28$ &  $0.69 \pm 0.23$ &  $0.81 
			\pm 0.34$ &  
			$0.94 \pm 0.40$ & $0.92 \pm 0.23$ &  $0.96 \pm 0.39$ &  $0.81 \pm 
			0.24$ &  
			$0.76 \pm 0.22$ &  
			$0.66 \pm 0.20$ \\
			\textsc{[300,310]} & $0.87 \pm 0.25$ &  $0.71 \pm 0.26$ &  $0.83 
			\pm 0.36$ &  
			$0.97 \pm 0.39$ &  
			$0.92 \pm 0.28$ &  $0.90 \pm 0.35$ &  $0.81 \pm 0.24$ &  $0.75 \pm 
			0.24$ &  
			$0.61 \pm 0.24$ \\
			\textsc{[350,360]} & $0.84 \pm 0.24$ &  $0.70 \pm 0.23$ &  $0.82 
			\pm 0.36$ &  
			$0.91 \pm 0.37$ &  
			$0.81 \pm 0.26$ &  $0.86 \pm 0.31$ &  $0.80 \pm 0.21$ &  $0.75 \pm 
			0.19$ &  
			$0.59 \pm 0.20$ \\
			\textsc{[400,410]} & $0.90 \pm 0.27$ &  $0.70 \pm 0.23$ &  $0.83 
			\pm 0.36$ &  
			$0.88 \pm 0.32$ &  
			$0.77 \pm 0.23$ &  $0.88 \pm 0.30$ &  $0.81 \pm 0.21$ &  $0.74 \pm 
			0.20$ &  
			$0.58 \pm 0.20$ \\
			\textsc{[450,460]} & $0.89 \pm 0.25$ &  $0.70 \pm 0.23$ &  $0.83 
			\pm 0.36$ &  
			$0.88 \pm 0.32$ &  
			$0.77 \pm 0.23$ &  $0.88 \pm 0.30$ &  $0.81 \pm 0.21$ &  $0.74 \pm 
			0.20$ &  
			$0.58 \pm 0.20$ \\
			\textsc{[500,510]} & $0.89 \pm 0.26$ &  $0.72 \pm 0.22$ &  $0.84 
			\pm 0.30$ &   
			$0.99 \pm 0.42$ & $0.92 \pm 0.24$ &  $0.95 \pm 0.46$ &  $0.81 \pm 
			0.23$ &  
			$0.80 \pm 0.25$ &  
			$0.61 \pm 0.20$ \\
			\textsc{[550,560]} & $0.88 \pm 0.26$ &  $0.72 \pm 0.24$ &  $0.89 
			\pm 0.42$ &  
			$0.95 \pm 0.37$ &  
			$0.86 \pm 0.27$ &  $0.90 \pm 0.28$ &  $0.81 \pm 0.20$ &  $0.78 \pm 
			0.23$ &  
			$0.63 \pm 0.21$ \\
			\hline
			\textsc{SCIP\_TUNED} & $0.89 \pm 0.28$ &  $0.77 \pm 0.23$ &  $0.99 
			\pm 0.31$ &  
			$1.08 \pm 0.45$ &  
			$1.05 \pm 0.28$ &  $1.03 \pm 0.38$ &  $0.94 \pm 0.23$ &  $0.91 \pm 
			0.28$ &  
			$0.76 \pm 0.25$ \\
			\hline
			\hline
		\end{tabular}
		\caption{Average relative primal integral (mean $\pm$ std.) of schedule 
		(with diving heuristics) w.r.t. 
		default 
			SCIP over GISP 
			instances derived from random graphs. The first nine rows 
			correspond to 
			schedules that were trained on instances 
			of size 
			\textsc{[l,u]}}
		\label{table:crossvalidation}
	\end{table}

\begin{table}[t]
	\renewcommand{\arraystretch}{1.35}
	\tiny
	\centering
	\begin{tabular}{lcc}
		\hline
		\hline
		schedule & better primal integral & better primal bound \\
		\hline
		\textsc{[150,160]} & 0.69 & 0.70 \\
		\textsc{[200,210]} & 0.69 & 0.65 \\
		\textsc{[250,260]} & 0.68 & 0.55 \\
		\textsc{[300,310]} & 0.72 & 0.58 \\
		\textsc{[350,360]} & 0.76 & 0.62 \\
		\textsc{[400,410]} & 0.75 & 0.61 \\
		\textsc{[450,460]} & 0.75 & 0.61 \\
		\textsc{[500,510]} & 0.68 & 0.58 \\
		\textsc{[550,560]} & 0.70 & 0.59 \\
		\hline
		\hline
	\end{tabular}
	\caption{Fraction of instances for which our method's schedule (with diving 
	heuristics) has a better 
	primal integral/bound at termination w.r.t. \textsc{scip\_tuned}. Only 
		instances that were not solved to optimality by both 
		\textsc{scip\_tuned} and the schedule are considered in the second 
		column.}
	\label{table:crossvalidation2}
\end{table}

	Over all test sets, the schedules terminated with a strictly better primal
	integral on 69--76\% and with a strictly better primal bound on 59--70\% 
	of the instances compared to \textsc{scip\_tuned} (see Table 
	\ref{table:crossvalidation2}).
	
	Table \ref{table:crossvalidation_LNS} shows a part of the 
	cross-validation experiments for schedules containing diving and LNS 
	heuristics. As expected, including both classes of heuristics improves the 
	overall performance of the schedule. In this case, the improvement is only 
	marginal since on the instances we consider, diving seems to perform 
	significantly better than LNS.
	
	\begin{table}[t]
		\renewcommand{\arraystretch}{1.35}
		\tiny
		\centering
		\begin{tabular}{lccccc}
			\hline
			\hline
			\diagbox[width=6em]{train}{test} & \textsc{[150,160]} & 
			\textsc{[200,210]} & 
			\textsc{[250,260]} & \textsc{[300,310]} & \textsc{[350,360]} \\
			\hline
			\textsc{[150,160]} & \hlfair{$0.84 \pm 0.19$} & \hlintense{$0.65 
			\pm 0.29$} & $0.89 \pm 0.35$ & \hlmediocre{$0.91 \pm 0.32$} &  
			$0.88 \pm 0.28$ \\
			\textsc{[200,210]} & \hlmediocre{$0.87 \pm 0.16$} &  $0.76 \pm 
			0.33$ & $0.91 \pm 0.32$ & $0.93 \pm 0.34$ & \hlmediocre{$0.89 \pm 
			0.29$} \\
			\textsc{[250,260]} & \hlfair{$0.83 \pm 0.18$} & $0.71 \pm 0.34$ &  
			$0.89 \pm 0.31$ & \hlmediocre{$0.90 \pm 0.29$} & \hlmediocre{$0.87 
			\pm 0.32$} \\
			\textsc{[300,310]} & \hlmediocre{$0.81 \pm 0.19$} & 
			\hlintense{$0.62 \pm 0.24$} & $0.91 \pm 0.42$ &  
			\hlmediocre{$0.92 \pm 0.32$} & $0.94 \pm 0.32$ \\
			\textsc{[350,360]} & \hlfair{$0.82 \pm 0.19$} & \hlintense{$0.61 
			\pm 0.25$} & $0.84 \pm 0.42$ & \hlmediocre{$0.86 \pm 0.23$} &  
			$0.86 \pm 0.28$ \\
			\hline
			\textsc{SCIP\_TUNED} & $0.89 \pm 0.28$ &  $0.77 \pm 0.23$ &  $0.99 
			\pm 
			0.31$ 
			&  
			$1.08 \pm 0.45$ &  $1.05 \pm 0.28$ \\	
			\hline
			\hline
		\end{tabular}
		\caption{Average relative primal integral (mean $\pm$ std.) of schedule 
		(with diving and LNS heuristics) w.r.t. 
		default 
			SCIP over GISP instances derived from random graphs. The first five 
			rows correspond to schedules that 
			were trained on instances of size \textsc{[l,u]}. On 
			highlighted entries, a schedule controling both diving and LNS 
			performs better than its diving counterpart (see Table 
			\ref{table:crossvalidation}). More intense colors denote higher 
			improvement.}
		\label{table:crossvalidation_LNS}
	\end{table}
	
	\textbf{Finding a schedule with SMAC.}
	As mentioned before, we can also 
	find a schedule by using the hyperparameter tuning tool SMAC. To test 
	SMAC's performance on the random graph instances, we trained ten SMAC 
	schedules on a selection of the nine training sets. To make it easier for 
	SMAC, we only considered diving heuristics in this case. For the sake of 
	comparability, we gave SMAC the same total computational time for training 
	as we did in data collection: With 25 training instances per set and a time 
	limit of four hours, this comes to 100 hours per training set and schedule. 
	Note that since SMAC runs sequentially, training the SMAC schedules took 
	over four days per schedule, whereas training a schedule following the 
	greedy algorithm only took four hours with enough machines. To pick the 
	best performing SMAC schedule for each training set, we ran all ten 
	schedules on the test set of same size as the corresponding training 
	set
	and chose the best performing one to also run on the other test sets.
	
	The results can be found in Table \ref{table:crossvalidation_SMAC}. As we 
	can see, on all test sets, all schedules are significantly better than 
	default SCIP. However, when comparing these results to the performance of 
	the greedy schedules (see Table \ref{table:crossvalidation}), we can see 
	that SMAC performs worse on average. Over all five test sets, the SMAC 
	schedules terminated with a strictly better primal integral on 36 -- 54\% 
	and with a strictly better primal bound on 37 -- 55\% of the instances.
	\begin{table}[t]
		\renewcommand{\arraystretch}{1.35}
		\tiny
		\centering
		\begin{tabular}{lccccc|cccc}
			\hline
			\hline
			\multirow{2}{*}{\diagbox[width=6em]{train}{test}} & 
			\multirow{2}{*}{\textsc{[150,160]}} & 
			\multirow{2}{*}{\textsc{[250,260]}} & 
			\multirow{2}{*}{\textsc{[350,360]}} & 
			\multirow{2}{*}{\textsc{[450,460]}} & 
			\multirow{2}{*}{\textsc{[550,560]}} && 
			\multicolumn{2}{c}{compared to \textsc{schedule}} &\\
			\cline{8-9} 
			& & & & & && better primal integral & better primal bound &\\
			\hline
			\textsc{[150,160]} & \hlintense{$0.81 \pm 0.23$} & 
			\hlintense{$0.77 \pm 
			0.34$} & $0.90 \pm 0.27$ & $0.85 \pm 0.24$ & $0.65 \pm 0.19$ && 
			0.49 
			& 0.37 &\\
			\textsc{[250,260]} & \hlfair{$0.87 \pm 0.26$} & $0.88 \pm 0.42$ & 
			\hlmediocre{$0.87 \pm 0.25$} & $0.83 \pm 0.24$ & \hlintense{$0.59 
			\pm 0.22$} && 0.52 & 0.53 &\\
			\textsc{[350,360]} & $0.86 \pm 0.24$ & \hlfair{$0.80 \pm 0.37$} & 
			$0.86 
			\pm 
			0.25$ & $0.80 \pm 0.24$ & $0.68 \pm 0.18$ && 0.47 & 
			0.42 &\\
			\textsc{[450,460]} & $0.93 \pm 0.26$ & $0.87 \pm 0.32$ & $0.90 \pm 
			0.19$ & $0.85 \pm 0.25$ & $0.69 \pm 0.23$ && 0.36 & 
			0.44 &\\
			\textsc{[550,560]} & \hlfair{$0.87 \pm 0.22$} & \hlmediocre{$0.83 
			\pm 0.31$} & 
			$0.92 \pm 
			0.29$ & $0.84 \pm 0.26$ & \hlmediocre{$0.58 \pm 0.21$} && 0.54 & 
			0.55 &\\
			\hline
			\textsc{scip\_tuned} & $0.89 \pm 0.28$ &  $0.77 \pm 0.23$ &  $0.99 
			\pm 
			0.31$ 
			&  
			$1.08 \pm 0.45$ &  $1.05 \pm 0.28$ && - & - &\\	
			\hline
			\hline
		\end{tabular}
		\caption{Average relative primal integral (mean $\pm$ std.) of SMAC 
		schedule w.r.t. default 
			SCIP and the fraction of instances for which the SMAC schedule 
			has a better primal integral/bound at termination w.r.t. the greedy 
			schedule over GISP instances derived from random graphs. The 
			first five rows correspond to schedules (with diving heuristics) 
			that were trained with SMAC on instances of size \textsc{[l,u]}. On 
			highlighted entries, a SMAC schedule performs better than its 
			greedy counterpart (see Table \ref{table:crossvalidation}). More 
			intense colors denote higher improvement. Only instances that were 
			not solved to optimality by both SMAC and the greedy schedule are 
			considered in the last column.}
		\label{table:crossvalidation_SMAC}
	\end{table}
	\textbf{DIMACS graph instances.} Table \ref{table:dimacs} summarizes the 
	results on the instances derived from DIMACS graphs. To stay consistent 
	with \cite{khalil17}, we only schedule diving heuristics. As we can see, 
	the schedule setting dominates default SCIP in all metrics, but an 
	especially drastic 
	improvement can be obtained w.r.t. the primal integral: the schedule 
	reduces the primal integral by 49\%.

   When looking at the total time spent in heuristics, 
	we see that heuristics run significantly shorter but with more success: On 
	average, the incumbent success rate is higher compared to
	default SCIP.
	
		\begin{table}
		\renewcommand{\arraystretch}{1.35}
		\scriptsize
		\centering
		\begin{tabular}{lS[table-format=4.2]S[table-format=4.2]S[table-format=4.2]}
			\hline
			\hline
			& \textsc{DEFAULT} & \textsc{SCIP\_TUNED} & \textsc{SCHEDULE} \\
			\hline
			Primal Integral & 934.48 & 555.75 & 470.73 \\
			Time to first Incumbent & 1.33 & 1.33 & 1.26 \\
			Time to best Incumbent & 4266.68 & 2642.46 & 2803.38 \\
			Best Incumbent  & 2382.03 & 2385.73 & 2404.63 \\
			\hline
			Total heuristic calls* & 138.57 & 137.38 & 140.03 \\ 
			Total heuristic time* & 258.88 & 304.96 & 190.10 \\
			Number of Incumbents found* & 2.72 & 3.08 & 3.33 \\
			Incumbent Success Rate* & 0.01 & 0.02 & 0.02 \\
			\hline
			Gap & 144.59 & 144.03 & 141.70 \\
			Primal-dual Integral & 450148.72 & 435321.67 & 430882.04 \\
			\hline
			\hline
		\end{tabular}
		\caption{Summary of results on GISP instances derived from DIMACS 
			graphs. Values shown are aggregates over instances; geometric means 
			are 
			used. Statictics with * refer only to the heuristics used in the 
			schedule.}
		\label{table:dimacs}
	\end{table}
	
	\section{Conclusion and Discussion}
	In this work, we propose a data-driven framework for scheduling primal 
	heuristics in a MIP solver such that the primal performance is optimized. 
	Central to our approach is a novel formulation of the learning task as a 
	scheduling problem, an efficient data collection procedure, and a fast, 
	effective heuristic for solving the learning problem on a training dataset. 
	A comprehensive experimental evaluation shows that our approach 
	consistently learns heuristic schedules with better primal performance than 
	SCIP's default settings. Furtheremore, by replacing our heuristic algorithm 
	with the hyperparameter tuning tool SMAC in our scheduling framework, we 
	are able to obtain a worse but still significant performance improvement 
	w.r.t. SCIP's default. Together with the prohibitive computational costs of SMAC, we conclude that for our heuristic scheduling 
	problem, the proposed heuristic algorithm constitutes an efficient alternative to 
	existing methods.
	
	A possible limitation of our approach is that it produces a single, ``one-size-fits-all" schedule for a class of training instances. It is thus natural to wonder whether alternative formulations of the learning problem that leverage additional contextual data about an input MIP instance and/or a heuristic can be useful. We note that learning a mapping from the space of MIP instances to the space of possible schedules is not trivial. The space of possible schedules is a highly structured output space that involves both the permutation over heuristics and their respective iteration limits. The approach proposed here is much simpler in nature, which makes it easy to implement and incorporate into a sophisticated MIP solver.
	
	Although we have framed the heuristic scheduling problem in machine 
	learning terms, we are yet to analyze the learning-theoretic aspects of the 
	problem. More specifically, our approach is justified on empirical grounds 
	in Section~\ref{sec:expresults}, but we are yet to attempt to analyze potential generalization guarantees. We view the recent foundational 
	results by \cite{balcan2019much} as a promising framework that may apply to 
	our setting, as it has been used for the branching problem in 
	MIP \cite{balcan2018learning}.
	
	\clearpage
	\bibliographystyle{apalike}
	\bibliography{arxiv_bib}

\begin{thebibliography}{}

\bibitem[Balcan et~al., 2019]{balcan2019much}
Balcan, M.-F., DeBlasio, D., Dick, T., Kingsford, C., Sandholm, T., and
  Vitercik, E. (2019).
\newblock How much data is sufficient to learn high-performing algorithms?
\newblock {\em arXiv preprint arXiv:1908.02894}.

\bibitem[Balcan et~al., 2018]{balcan2018learning}
Balcan, M.-F., Dick, T., Sandholm, T., and Vitercik, E. (2018).
\newblock Learning to branch.
\newblock In {\em International conference on machine learning}, pages
  344--353. PMLR.

\bibitem[Baltean-Lugojan et~al., 2019]{baltean19}
Baltean-Lugojan, R., Bonami, P., Misener, R., and Tramontani, A. (2019).
\newblock Scoring positive semidefinite cutting planes for quadratic
  optimization via trained neural networks.

\bibitem[Berthold, 2006]{berthold06}
Berthold, T. (2006).
\newblock Primal heuristics for mixed integer programs.
\newblock Master's thesis.

\bibitem[Berthold, 2013a]{berthold132}
Berthold, T. (2013a).
\newblock Measuring the impact of primal heuristics.
\newblock {\em Operations Research Letters}, 41(6):611 -- 614.

\bibitem[Berthold, 2013b]{berthold13}
Berthold, T. (2013b).
\newblock Primal minlp heuristics in a nutshell.
\newblock In {\em OR}.

\bibitem[Berthold, 2018]{berthold18}
Berthold, T. (2018).
\newblock A computational study of primal heuristics inside an mi(nl)p solver.
\newblock {\em Journal of Global Optimization}, 70:189--206.

\bibitem[Colombi et~al., 2017]{colombi17}
Colombi, M., Mansini, R., and Savelsbergh, M. (2017).
\newblock The generalized independent set problem: Polyhedral analysis and
  solution approaches.
\newblock {\em European Journal of Operational Research}, 260:41--55.

\bibitem[Gamrath et~al., 2020]{gamrath20}
Gamrath, G., Anderson, D., Bestuzheva, K., Chen, W.-K., Eifler, L., Gasse, M.,
  Gemander, P., Gleixner, A., Gottwald, L., Halbig, K., Hendel, G., Hojny, C.,
  Koch, T., Bodic, P.~L., Maher, S.~J., Matter, F., Miltenberger, M.,
  M{\"u}hmer, E., M{\"u}ller, B., Pfetsch, M., Schl{\"o}sser, F., Serrano, F.,
  Shinano, Y., Tawfik, C., Vigerske, S., Wegscheider, F., Weninger, D., and
  Witzig, J. (2020).
\newblock {The SCIP Optimization Suite 7.0}.
\newblock ZIB-Report 20-10, Zuse Institute Berlin.

\bibitem[He et~al., 2014]{he14}
He, H., III, H.~D., and Eisner, J.~M. (2014).
\newblock Learning to search in branch and bound algorithms.
\newblock In {\em Advances in Neural Information Processing Systems},
  volume~27, pages 3293--3301. Curran Associates, Inc.

\bibitem[Hendel, 2018]{hendel18}
Hendel, G. (2018).
\newblock Adaptive large neighborhood search for mixed integer programming.
\newblock {\em Mathematical Programming Computation}.
\newblock under review.

\bibitem[Hendel et~al., 2018]{hendel182}
Hendel, G., Miltenberger, M., and Witzig, J. (2018).
\newblock Adaptive algorithmic behavior for solving mixed integer programs
  using bandit algorithms.
\newblock In {\em OR 2018: International Conference on Operations Research}.

\bibitem[Hewitt et~al., 2010]{hewitt10}
Hewitt, M., Nemhauser, G., and Savelsbergh, M. (2010).
\newblock Combining exact and heuristic approaches for the capacitated
  fixed-charge network flow problem.
\newblock {\em INFORMS Journal on Computing}, 22:314--325.

\bibitem[Hochbaum and Pathria, 1997]{hochbaum97}
Hochbaum, D.~S. and Pathria, A. (1997).
\newblock Forest harvesting and minimum cuts: A new approach to handling
  spatial constraints.
\newblock {\em Forest Science}, 43:544--554.

\bibitem[Hutter et~al., 2009]{hutter09}
Hutter, F., Hoos, H., Leyton-Brown, K., and St{\"u}tzle, T. (2009).
\newblock Paramils: An automatic algorithm configuration framework.
\newblock {\em J. Artif. Intell. Res. (JAIR)}, 36:267--306.

\bibitem[Hutter et~al., 2011]{HutterHoosLeytonbrown2011}
Hutter, F., Hoos, H.~H., and Leyton-Brown, K. (2011).
\newblock Sequential model-based optimization for general algorithm
  configuration.
\newblock In Coello, C. A.~C., editor, {\em Learning and Intelligent
  Optimization}, pages 507--523, Berlin, Heidelberg. Springer Berlin
  Heidelberg.

\bibitem[Khalil et~al., 2016]{khalil16}
Khalil, E.~B., Bodic, P.~L., Song, L., Nemhauser, G., and Dilkina, B. (2016).
\newblock Learning to branch in mixed integer programming.
\newblock In {\em Proceedings of the 30th AAAI Conference on Artificial
  Intelligence}.

\bibitem[Khalil et~al., 2017]{khalil17}
Khalil, E.~B., Dilkina, B., Nemhauser, G., Ahmed, S., and Shao, Y. (2017).
\newblock Learning to run heuristics in tree search.
\newblock In {\em 26th International Joint Conference on Artificial
  Intelligence (IJCAI)}, pages 659--666.

\bibitem[Kruber et~al., 2017]{kruber17}
Kruber, M., L{\"u}bbecke, M., and Parmentier, A. (2017).
\newblock Learning when to use a decomposition.
\newblock In {\em Lecture Notes in Computer Science}, pages 202--210.

\bibitem[Lodi, 2013]{lodi132}
Lodi, A. (2013).
\newblock The heuristic (dark) side of mip solvers.
\newblock {\em Hybrid Metaheuristics}, 434:273--284.

\bibitem[Lodi and Tramontani, 2013]{lodi13}
Lodi, A. and Tramontani, A. (2013).
\newblock Performance variability in mixed-integer programming.
\newblock {\em Tutorials in Operations Research, Vol. 10}, pages 1--12.

\bibitem[Munagala et~al., 2005]{munagala05}
Munagala, K., Babu, S., Motwani, R., and Widom, J. (2005).
\newblock The pipelined set cover problem.
\newblock In {\em International Conference on Database Theory}, volume 3363,
  pages 83--98.

\bibitem[Nair et~al., 2020]{nair20}
Nair, V., Bartunov, S., Gimeno, F., von Glehn, I., Lichocki, P., Lobov, I.,
  O'Donoghue, B., Sonnerat, N., Tjandraatmadja, C., Wang, P., Addanki, R.,
  Hapuarachchi, T., Keck, T., Keeling, J., Kohli, P., Ktena, I., Li, Y.,
  Vinyals, O., and Zwols, Y. (2020).
\newblock Solving mixed integer programs using neural networks.

\bibitem[Streeter, 2007]{streeter07}
Streeter, M. (2007).
\newblock {\em Using Online Algorithms to Solve NP-Hard Problems More
  Efficiently in Practice}.
\newblock PhD thesis, Carnegie Mellon University.

\end{thebibliography}
	
    \clearpage
	\appendix
	\section{Formulating the Scheduling Problem as a MIQP}
	\label{sec:mip}
	
	In this section, we present the exact formulation of Problem 
	\eqref{eq:schedulingproblem} as a MIQP. First, we describe the parameters 
	and variables we need to formulate the problem. Then, we state the problem 
	and explain shortly what every constraint represents.
	
	\subsection{Parameters}
	\begin{itemize}
		\item[--] $\mathcal{D}$, set of data points coming from a set of MIP 
		instances $\mathcal{X}$. Each data point is of the form $(h, N, 
		\tau^h_N)$, where $h \in \mathcal{H}$ is a heuristic, 
		$N\in\mathcal{N}_{\mathcal{X}}$ 
		indexes a node of the B\&B tree of $\mathcal{X}$, and $\tau^h_N$ is the 
		number of iterations $h$ needed to find a solution to $N$ (if $h$ could 
		not solve $N$, we set $\tau^h_N = \infty$).
		
		\item[--] $T^h, \forall h \in \mathcal{H}$, the maximum number of 
		iterations $h$ needed to find a solution, i.e., $T^h \coloneqq 
		\text{max}\{ 
		\tau_N^h < \infty \mid N \in \mathcal{N}_{\mathcal{X}}\}$. 
		
		\item[--] $\alpha \in [0,1]$, minimal fraction of nodes the resulting 
		schedule should solve.
	\end{itemize}
	
	\subsection{Variables Describing the Schedule}
	\begin{itemize}
		\item[--] $x^h_p \in \{0,1\}, \forall h \in \mathcal{H}, \forall p \in 
		\{0, 
		\dots, |\mathcal{H}| \}$: The variable is set to 1 if $h$ is executed 
		at position $p$ in the schedule (if $x^h_0 = 1$ then $h$ is not in the 
		schedule).
		
		\item[--] $t^h \in [0, \dots, T^h], \forall h \in \mathcal{H}$: This 
		integer variable is equal to the maximal number of iterations $h$ can 
		use in the schedule (we set $t^h = 0$ if $h$ is not in the schedule).
	\end{itemize}
	
	\subsection{Auxiliary Variables}
	\begin{itemize}
		\item[--] $p^h \in \{0, \dots, |\mathcal{H}|\}, \forall h \in 
		\mathcal{H}$: 
		This integer variable is equal to the position of $h$ in the schedule.
		
		\item[--] $s^h_N \in \{0,1\}, \forall h \in \mathcal{H}, \forall N \in 
		\mathcal{N}_{\mathcal{X}}$: The variable is set to 1 if heuristic $h$ 
		solves node $N$ in the schedule.
		
		\item[--] $s_N \in \{0,1\}, \forall N \in \mathcal{N}_{\mathcal{X}}$: 
		This 
		variable is set to 1 if the schedule solves node $N$.
		
		\item[--] $p_N^{min} \in \{1, \dots, |\mathcal{H}|\}, \forall N \in 
		\mathcal{N}_{\mathcal{X}}$: This integer variable is equal to the 
		position of the 
		heuristic that first solves node $N$ in the schedule (if the schedule 
		does not solve $N$, we set it to $|\mathcal{H}|$).
		
		\item[--] $z_N^h \in \{0,1\}, \forall h \in \mathcal{H}, \forall N \in 
		\mathcal{N}_{\mathcal{X}}$: This variable is set to 1 if $h$ is 
		executed before position $p_N^{min}$.
		
		\item[--] $f_N^h \in \{0,1\}, \forall h \in \mathcal{H}, \forall N \in 
		\mathcal{N}_{\mathcal{X}}$: The variable is set to 1 if $h$ is the 
		heuristic that solves $N$ first, i.e., if $p^h = p_N^{min}$.
		
		\item[--] $t_N \in \{1, \dots, 1 + \sum_{h} T^h \}, 
		\forall N \in \mathcal{N}_{\mathcal{X}}$: This integer variable is 
		equal to the total number of iterations the schedule needs to solve 
		node $N$ (if $N$ is not solved by the schedule, we set it to 1 plus the 
		total length of the schedule, i.e., $1 + \sum_{p} \sum_h x_p^h t^h$).
	\end{itemize}
	
	\subsection{Formulation}
	\label{sec:MIQP}
	In the following, we give an explicit formulation of 
	\eqref{eq:schedulingproblem} as a MIQP. Note that some constraints use 
	nonlinear functions like the maximum/minimum of a finite set or the 
	indicator function 
	$\mathbbm{1}$. These can be easily linearized by introducing additional 
	variables and constraints, thus the following formulation is indeed a MIQP. 
	For the sake of readability, we omit stating all the linearizations 
	explicitly.
	
	\setcounter{equation}{0}
	\begin{align}
		\min 
		\label{eq:objective}\quad & \sum_{N} t_N \\
		\textrm{s.t.}
		\label{eq:posunique}\quad & \sum_h x_p^h \leq 1, \forall p \in \{1, 
		\dots, |\mathcal{H}|\} \\
		\label{eq:varposunique}\quad & \sum_{p} x_p^h = 1, \forall h \in 
		\mathcal{H} \\
		\label{eq:pos}\quad & p^h = \sum_{p} p x^h_p,  \forall h \in 
		\mathcal{H}\\
		\label{eq:timebound}\quad & T^h(1-x^h_0) \geq t^h, \forall h \in 
		\mathcal{H}\\
		\label{eq:hsolvesn2} \quad & s_N^h = \text{max} \{0, \min\{1, t^h - 
		\tau_N^h + 1\}\}, \forall h \in \mathcal{H}, \forall N \in 
		\mathcal{N}_{\mathcal{X}} \\
		\label{eq:nsolved}\quad & s_N = \text{min} \{1, \sum_h s_N^h\}, \forall 
		N \in \mathcal{N}_{\mathcal{X}}\\
		\label{eq:solvesenough}\quad & \frac{1}{|\mathcal{N}_{\mathcal{X}}|} 
		\sum_{N} s_N \geq \alpha \\
		\label{eq:minpos} \quad & p_N^{min} = \text{min}\{ p^h s_N^h + (1 - 
		s_N^h) \mid 
		\mathcal{H}|) \mid h \in \mathcal{H} \}, \forall N \in 
		\mathcal{N}_{\mathcal{X}}\\
		\label{eq:posbeforemin}\quad & z_N^h = \mathbbm{1}_{\{p^h < 
		p_N^{min}\}},\forall h \in \mathcal{H}, \forall N \in 
		\mathcal{N}_{\mathcal{X}} \\
		\label{eq:posequalmin}\quad & f_N^h = \mathbbm{1}_{\{p^h = 
		p_N^{min}\}}, \forall h \in \mathcal{H}, \forall N \in 
		\mathcal{N}_{\mathcal{X}} \\
		\label{eq:ntime} \quad & t_N = s_N( \sum_h z_N^h t^h + f_N^h \tau_N^h) 
		+ (1 - s_n) ( 1 + \sum_{h,p} x_p^h t^h ), \forall N \in 
		\mathcal{N}_{\mathcal{X}}
	\end{align}
	
	\eqref{eq:objective} calculates the total number of iterations the schedule 
	needs to solve all nodes.
	
	\eqref{eq:posunique} and \eqref{eq:varposunique} guarantee that only one 
	copy of each heuristic is run, and that every non-zero position is occupied 
	by at most one heuristic.
	
	\eqref{eq:pos} calculates the position of a heuristic in the schedule.
	
	\eqref{eq:timebound} ensures that $t^h = 0$ if $h$ is not in the schedule.
	
	\eqref{eq:hsolvesn2} forces $s_N^h$ to 1 if $h$ solves node $N$ in the 
	schedule.
	
	\eqref{eq:nsolved} forces $s_N$ to 1 if the schedule solves node $N$.
	
	\eqref{eq:solvesenough} guarantees that the schedules solves enough nodes.
	
	\eqref{eq:minpos} calculates the position of the first heuristic that 
	solves $N$ in the schedule.
	
	\eqref{eq:posbeforemin} forces $z_N^h$ to 1 if $h$ is executed before 
	position $p_N^{min}$.
	
	\eqref{eq:posequalmin} forces $f_N^h$ to 1 if $h$ is executed at position 
	$p_N^{min}$.
	
	\eqref{eq:ntime} calculates the number of iterations necessary for the 
	schedule to solve $N$.
	
	\section{Implementation Details}
	\label{sec:implementation}
	Not every idea that works in theory can be directly translated to also work 
	in practice. Hence, it is sometimes inevitable to adapt ideas and make 
	compromises when implementing a new method. In this section, we touch upon 
	 aspects we needed to consider to ensure a reliable implementation of 
	the framework proposed in this paper.
	
	\textbf{Time as a measure of duration.} A heuristic schedule 
	controls two general aspects: The order and the duration for which the 
	different heuristics are executed. Even though it might seem intuitive to 
	use time to control the duration of a heuristic run, we use a suitable 
	proxy measure for every class of heuristics instead, as discussed 
	in Section \ref{sec:formulation}. There are two main problems that hinder 
	us from directly controlling time. 
	First, time is generally not stable enough to use for decision-making 
	within an optimization solver. To make it somewhat reliable, we would need 
	to solve instances exclusively on a single machine at a time. Hence, it 
	would not be possible to run instances in parallel which would cause the 
	solving process to be very expensive in practice. The second, even more 
	important problem is the following. Since the behavior of heuristics 
	significantly depends on different parameters, allowing the heuristic to 
	run for a longer time does not necessarily translate to a increase in 
	success probability if 
	crucial parameters are set to be very restrictive by default. That is why 
	we use a suitable proxy measure 
	for time instead of time itself.
	
	\textbf{Limitations of the shadow mode.} To make sure we obtain data 
	that is as independent as possible, the heuristics run in shadow mode 
	during data collection. This setting aims to 
	ensure that the heuristics only run in the background and do not report 
	anything back to the solver. However, it is not possible to
	hide \textit{all} of the heuristic's actions from SCIP. Since SCIP is not 
	designed to have heuristics running in the background, it is almost 
	impossible to locate and adjust the lines of code that influence the 
	solving process globally without limiting the heuristic's behavior too 
	much. For instance, one way of hiding all actions of diving heuristics 
	would be turning off propagation while diving. Since this would influence 
	the performance of the heuristics considerably, the resulting data would 
	not represent how the heuristics behave in practice. That is why we settled 
	with a shadow mode that hides most (and the most influential) of the 
	heuristic's activities from SCIP. 
\end{document}